\documentclass[10pt,twocolumn,letterpaper]{article}

\usepackage[hang,flushmargin]{footmisc}
\usepackage{cvpr}
\usepackage{times}
\usepackage{epsfig}
\usepackage{graphicx}
\usepackage{amsmath}
\usepackage{amssymb}

\usepackage{subfigure}
\usepackage{tabularx}
\usepackage{url}
\newcommand{\minisection}[1]{\vspace{0.04in} \noindent {\bf #1}\ \ }

\usepackage{float}
\usepackage{algorithm}

\usepackage[pagebackref=true,breaklinks=true,letterpaper=true,colorlinks,bookmarks=false]{hyperref}

\cvprfinalcopy 

\def\cvprPaperID{3180} 

\ifcvprfinal\fi

\begin{document}

\title{Leveraging Unlabeled Data for Crowd Counting by Learning to Rank}

\author{Xialei Liu\\
Computer Vision Center \\
Barcelona, Spain\\
{\tt\small xialei@cvc.uab.es}
\and
Joost van de Weijer\\
Computer Vision Center\\
Barcelona, Spain\\
{\tt\small joost@cvc.uab.es}
\and
Andrew D. Bagdanov\\
MICC, University of Florence \\
Florence, Italy\\
{\tt\small andrew.bagdanov@unifi.it}
}

\maketitle

\begin{abstract}
  We propose a novel crowd counting approach that
  leverages abundantly available unlabeled crowd imagery in a
  learning-to-rank framework. To induce a ranking of cropped images
, we use the  observation that any sub-image of a crowded scene image is guaranteed to contain the same number or fewer persons than the
  super-image. This allows us to address the problem of limited size
  of existing datasets for crowd counting.  We collect two crowd scene
  datasets from Google using keyword searches and query-by-example
  image retrieval, respectively. We demonstrate how to efficiently
  learn from these unlabeled datasets by incorporating
  learning-to-rank in a multi-task network which simultaneously ranks
  images and estimates crowd density maps.  Experiments on two of the
  most challenging crowd counting datasets show that our approach
  obtains state-of-the-art results.
\end{abstract}

\section{Introduction}

Crowd counting and crowd density estimation techniques aim to count
the number of persons in crowded scenes. They are essential in video
surveillance~\cite{chan2008privacy}, safety monitoring, and behavior analysis~\cite{sheng2016crowd}. Person counting and density estimation are instances of a broader class of classical counting problems in computer vision. Counting semantic image features is important in medical and biological image processing~\cite{lempitsky2010learning}, vehicle counting~\cite{onoro2016towards}, and numerous other application contexts.

\begin{figure}[t]
\centering
\includegraphics[width=0.9\columnwidth]{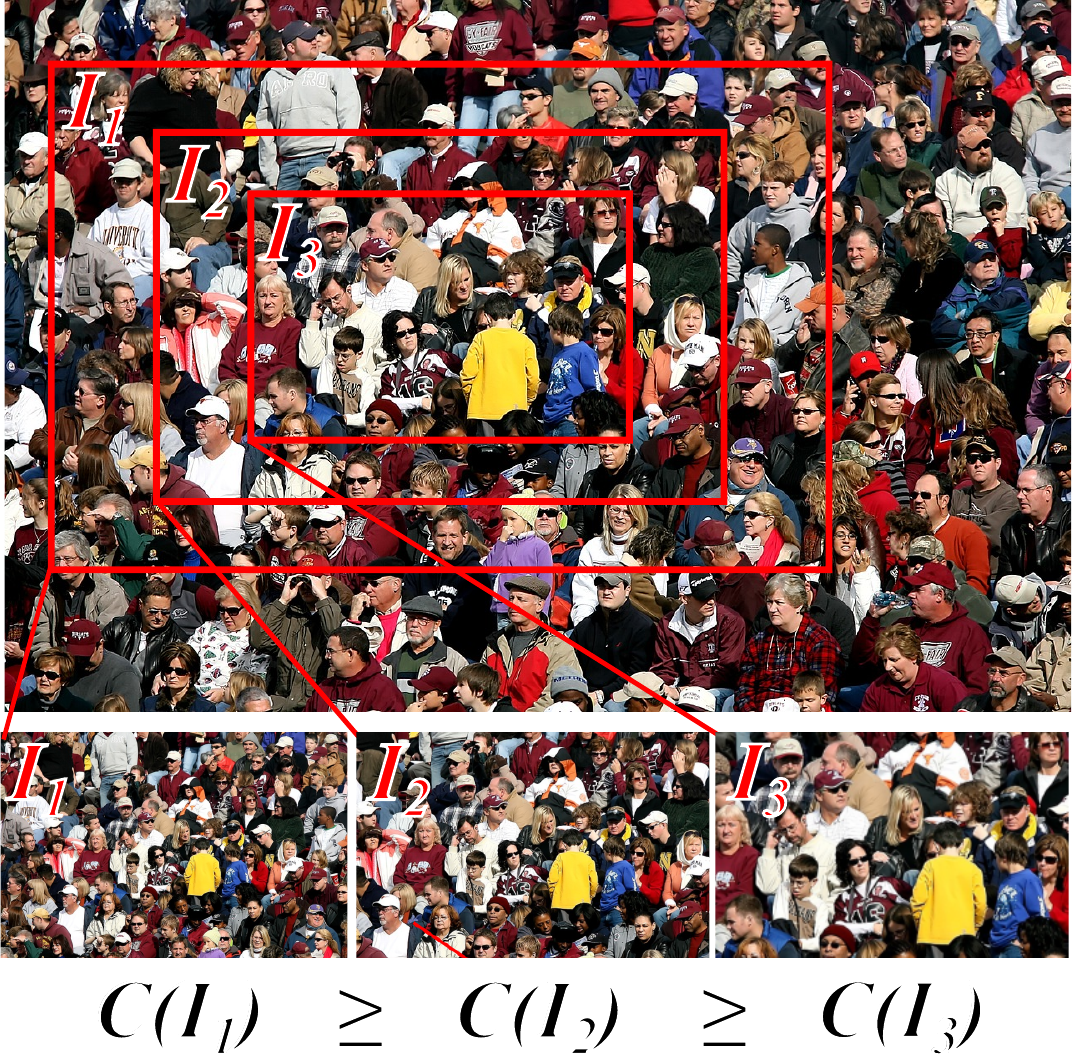}
\caption{Using ranked sub-images for self-supervised training.
  We sample a decreasing sequence of
  sub-images $I_1, I_2$, and $I_3$ from an \emph{unlabeled}
  image. Though we do not know the exact
  person counts $C(I_i)$, we use the fact that $C(I_1) \geq C(I_2)
  \geq C(I_3)$ as self-supervision to learn representations for person counting.}
\label{fig:overview}
\end{figure}

\begin{table*}[h]
\centering
\begin{tabular}{r|ccccc}
\hline
\textbf{Methods} & \textbf{Basic CNNs} & \textbf{Scale-aware} & \textbf{Context-aware} & \textbf{Multi-task}  & \textbf{Fast inference} \\ \hline\hline
Zhang et al. (2015)    \cite{zhang2015cross} & & & & \checkmark & \\
Shang et al.    \cite{shang2016end} & & & \checkmark & & \\
Marsden et al.  \cite{marsden2016fully} & \checkmark & & & & \checkmark \\
Zhang et al. (2016)   \cite{zhang2016single} &  & \checkmark  &  &  & \\
Babu Sam et al. \cite{Sam_2017_CVPR}      &     &  \checkmark    &            &        &           \\
Sindagi et al.  \cite{sindagi2017generating}   &            &   \checkmark  & \checkmark &         &       \\ \hline
Ours    & \checkmark & \checkmark    &     & \checkmark            & \checkmark     \\ \hline
\end{tabular}
\caption{State-of-the-art crowd counting networks and their characteristics. }
\label{table:property}
\end{table*}

Despite the attention the crowd counting problem has received, both classically and in the recent computer vision literature, it remains a difficult task in practice. Perspective distortion, clutter, occlusion, non-uniform distribution of people, complex illumination, scale variation, and a host of other scene-incidental imaging conditions render person counting and crowd density estimation in unconstrained images an extremely daunting problem. Techniques for crowd counting have been recently improved using
Convolutional Neural Networks (CNNs). These recent approaches include scale-aware regression models~\cite{onoro2016towards}, multi-column CNNs~\cite{zhang2016single}, and switching networks~\cite{Sam_2017_CVPR}. As with most CNN architectures, however, these person counting and crowd density estimation techniques are highly data-driven. Even modestly deep architectures for visual recognition require massive amounts of labeled training data for learning. For person counting, the labeling burden is even more onerous than usual. Training data for person counting requires that each individual person be meticulously labeled in training images. It is for this reason that person counting and crowd density estimation datasets tend to have only a few hundred images available for training. As a consequence, the ability to train these sophisticated CNN-based models suffers.

Recently, self-supervised learning has received more attention because
it provides an alternative to collecting large hand-labeled
datasets. Self-supervised learning is based on the idea of using an
auxiliary task (different, but related to the original supervised
task) for which data is freely available and no annotation is
required. As a consequence, self-supervised learning can be much more
scalable and flexible. A network trained to estimate the relative
location of patches in images was shown to automatically learn
features discriminative for semantic concepts
in~\cite{doersch2015unsupervised}. Other examples include methods that
can generate color images from gray scale images and vice
versa~\cite{larsson2017colorization,zhang2016colorful}, recover a
whole patch from the surrounding pixels by
inpainting~\cite{pathak2016context}, and learn from equivalence
relations~\cite{noroozi2017representation}.

In this paper, we propose a self-supervised task to improve the
training of networks for crowd counting.  Our approach leverages
\emph{unlabeled} crowd images at training time to significantly
improve performance. Our key insight is that even though we do not
have an \emph{exact} count of the number of persons in a crowd image,
we do know that crops sampled from a crowd image are guaranteed to
contain the same or fewer persons than the original (see
Figure~\ref{fig:overview}). This gives a technique for generating a
\emph{ranking} of sub-images that can be used to train a network to
estimate whether one image contains more persons than another
image. The standard approach to exploiting self-supervised learning is
to train the self-supervised task first, after which the resulting
network is fine-tuned on the final task for which limited data is
available. We show that this approach, which is used by the vast
majority of self-supervised
methods~\cite{doersch2015unsupervised,liu2017rankiqa,noroozi2017representation,pathak2016context,zhang2016colorful},
does not produce satisfactory results for crowd
counting. Our proposed self-supervision, however, yields significant
improvement over the state-of-the-art when added as a proxy task to
supervised crowd counting in a multi-task network.

The main contribution of this work is that we propose a method that
can leverage \emph{unlabeled} crowd imagery at training time. We
propose two different approaches to automatically acquire this data
from the Internet. In addition, we analyze three approaches to
training using ranked image sets in combination with datasets of
labeled crowd scenes. Finally, we demonstrate that our approach leads
to state-of-the-art results on two crowd counting datasets and obtains
excellent results on a cross-dataset experiment.

The rest of this paper is organized as follows. In the next section we
briefly review the literature related to crowd counting. Then, in
Section~\ref{sec:generating} we describe how to systematically
generate ranked images from unlabeled crowd imagery. In
Section~\ref{sec:method} we introduce our approach to exploiting this
ranked imagery at training time. We follow in
Section~\ref{sec:experiments} with an extensive experimental
evaluation of our approach and a comparative analysis with the
state-of-the-art.

\section{Related work}
We divide our discussion of related work into two main groups as
in~\cite{sindagi2017survey}: traditional approaches and CNN-based
methods. We focus on crowd counting in still images, but we refer the
interested reader to the following papers for examples of
crowd counting in
video~\cite{chan2012counting,cong2009flow,ma2013crossing}.

\begin{figure*}[tpb]
\centering
\subfigure{\includegraphics[width=0.95\textwidth]{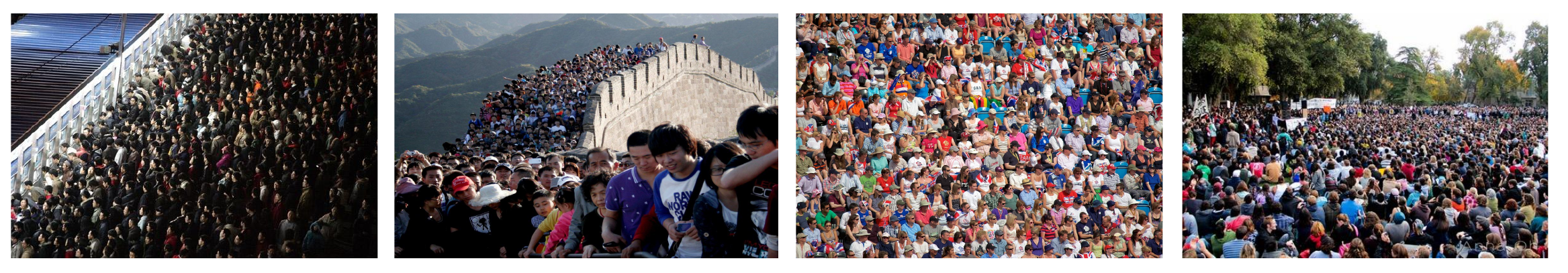}}
\subfigure{\includegraphics[width=0.95\textwidth]{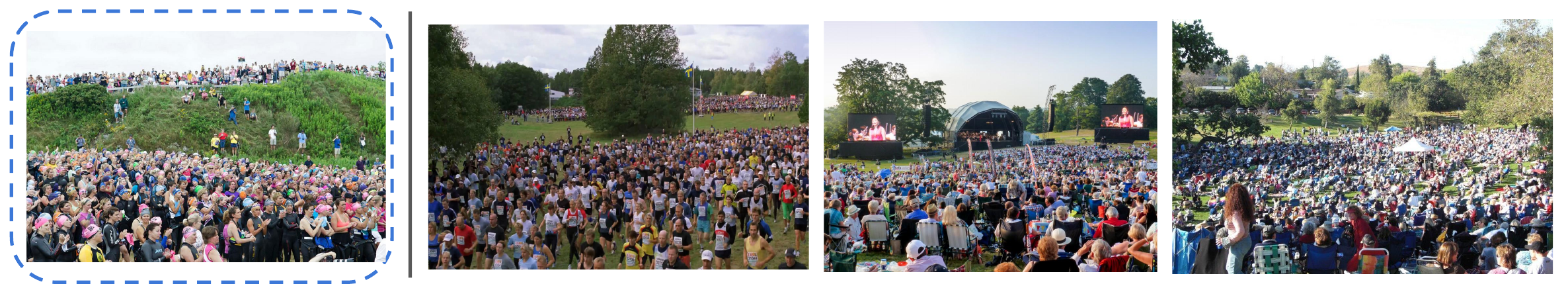}}
\caption{Example images from the retrieved crowd scene dataset. (top)
  Representative images using key words as query. (bottom)
  Representative images using training image as query image (the query
  image is depicted on the left).}
  \label{fig:data1}
\end{figure*}

Various traditional approaches have been proposed to deal with the
crowd counting problem. The main strategies are divided into the
various categories as in~\cite{loy2013crowd}. Most early work on crowd
counting used detectors to detect the heads or full bodies of persons
in the scene~\cite{dollar2012pedestrian,li2008estimating}. This
information can then be used to count.  However, detection-based
approaches fail in extremely dense crowded scenes due to occlusion and
low resolution of persons. To address these issues, researchers
proposed to map features learned from the crowded scene or patches to
the number of people~\cite{chan2009bayesian,chen2012feature}. By
counting using regression, the crowd counting problem is decomposed
into two parts: feature extraction and a regression model. While
regression-based approaches resulted in improvement, only global
counting was considered without any spatial information (i.e. without
estimating a density map). The authors of~\cite{lempitsky2010learning}
proposed to learn a mapping from  patches to corresponding
density maps, which achieved great success on a variety of counting
problems.

As introduced in the review of~\cite{sindagi2017survey}, CNN-based
approaches can be classified into different categories based on the
properties of the CNN (see Table~\ref{table:property} for an overview
of state-of-the-art networks and the properties they hold). Basic CNNs
incorporate only basic CNN layers in their networks. The approaches
in~\cite{fu2015fast,wang2015deep} use the AlexNet
network~\cite{krizhevsky2012imagenet} to map from crowd scene patches
to global number of people by changing the output of AlexNet from 1000
to 1. The resulting network can be trained end-to-end.

Due to the large variations of density in different images, recent
methods have focused on scale-awareness. The method proposed in
\cite{zhang2016single} trains a multi-column based architecture (MCNN)
to capture the different densities by using different sizes of kernels
in the network. Similarly, the authors of~\cite{onoro2016towards}
propose the Hydra-CNN architecture that takes different resolutions of
patches as inputs and has multiple output layers (heads) which are
combined in the end. Most recently, in~\cite{Sam_2017_CVPR} the
authors propose a switching CNN that can select an optimal head
instead of combining the information from all network heads. Finally,
context-aware models are networks that can learn from the context of
images. In~\cite{fu2015fast,sindagi2017generating} the authors propose
to classify images or patches into one of five classes: very high
density, high density, medium density, low density and very low
density. However, the definition of these five classes varies across
datasets and must be carefully chosen using knowledge of the
statistics of each dataset.

Although CNN-based methods have achieved great success in crowd
counting, due to lack of labeled data it is still challenging to train
deep CNNs without over-fitting. The authors of~\cite{zhang2015cross}
propose to learn density map and global counting in an alternating
sequence to obtain better local optima. The method
in~\cite{kang2016crowd} uses side information like ground-truth camera
angle and height to help the network to learn. However, this side
information is expensive to obtain and is not available in most
existing crowd counting datasets.

There are several works which have studied how to learn to rank, and
they focus on learning a ranking function from ground-truth
rankings~\cite{chen2009ranking,test1}. However, these are very
different from our approach in which we aim to learn \emph{from}
rankings. Most related to our work is a recent
paper~\cite{liu2017rankiqa} proposing a method for image quality
assessment using automatically generated rankings of distorted
images. In that work the authors used ranking data to pre-train a
network and then fine-tune on available labeled datasets.  We will
show that such an approach fails for crowd counting, and that only when
posed as a multi-task learning problem is the network able to exploit
the additional data from rankings.

In another recent paper~\cite{noroozi2017representation} a method is
proposed where the self-supervised task is to learn to count. The
authors propose two pretext tasks (scaling and tiling) which guide
self-supervised training. The network aims to learn to count visual
primitives in image regions. It is self-supervised by the fact that
the number of visual primitives is not expected to change under
scaling, and that the sum of all visual primitives in individual tiles
should equal the total number of visual primitives in the whole
image. Unlike our approach, they do not consider rankings of regions
and their counts are typically very low (several image
primitives). Also, their final tasks do not involve counting but
rather unsupervised learning of features for object recognition.

\minisection{Contributions with respect to the state-of-the-art:}
Basic CNNs are simple and fast to train, but usually achieve lower
accuracy. Combining different scale-aware models and context-aware
models has been shown to significantly increase performance, but
the complexity of these models is high. In addition, considering the
scarcity of large annotated datasets, ranked patches are used as side
information to decrease the effect of over-fitting. The model we
propose in this paper is fast to train, uses no side information,
supports fast inference, is scale-aware, is multi-task, and
outperforms the state-of-the-art. Our key contribution is in showing
how to effectively exploit unlabeled crowd imagery for pre-training
CNNs. In Table~\ref{table:property} we list current state-of-the-art
approaches to crowd counting along with ours and indicate the
characteristics of each.

\section{Generating ranked image sets for counting}\label{sec:generating}
As discussed in the introduction, acquiring data for crowd counting is
laborious because images often contain hundreds of persons which
require precise annotation. Instead, we propose a self-supervised task
for crowd-counting which exploits crowd images which are not
hand-labeled with person counts during training. Rather than
regressing to the absolute number of persons in the image, we train a
network which compares images and ranks them according to the number
of persons in the images. In this section, we show how to cheaply
collect rank-labeled data which can be used to train these methods.

The main idea is based on the observation that all patches contained within a larger patch must have a fewer or equal number of persons than the larger patch (see Figure~\ref{fig:overview}). This observation allows us to collect large datasets of crowd images for which relative ranks exist. Rather than having to painstakingly annotate each person we are only required to verify if the image contains a crowd. Given a crowd image we extract ranked patches according to Algorithm~\ref{table:algorithm2}. 

To collect a large dataset of crowd images from the Internet, we use
two different approaches:
\begin{itemize}
\item \textbf{Keyword query:} We collect a crowd scene dataset from
  Google Images by using different key words: \emph{Crowded,
    Demonstration, Train station, Mall, Studio, Beach}, all of which
  have high likelihood of containing a crowd scene. Then we delete
  images not relevant to our problem. In the end, we collected a
  dataset containing 1180 high resolution crowd scene images, which is
  about 24x the size of the UCF\_CC\_50 dataset, 2.5x the size of
  ShanghaiTech Part\_A, and 2x the size of ShanghaiTech Part\_B. Note
  that \emph{no other annotation of images is performed.} Example
  images from this dataset are given in Figure~\ref{fig:data1} (top
  row).

\item \textbf{Query-by-example image retrieval:} For each specific
  existing crowd counting dataset, we collect a dataset by using the
  training images as queries with the visual image search engine
  \emph{Google Images}. We choose the first ten similar images and
  remove irrelevant ones. For UCF\_CC\_50 we collect 256 images, for
  ShanghaiTech Part\_A 2229 images, and for ShanghaiTech Part\_B 3819
  images.  An example of images returned for a specific query image is
  given in Figure~\ref{fig:data1} (bottom row).
\end{itemize}

\begin{figure*}[tpb]
\centering
  \includegraphics[width=0.9\textwidth]{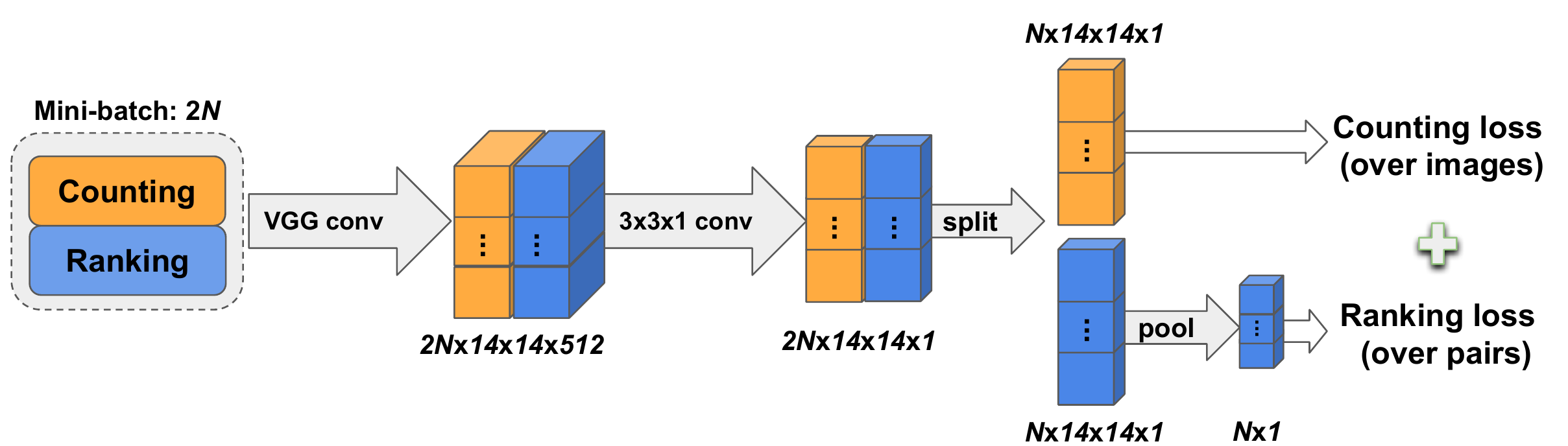}
  \caption{The multi-task framework combining both counting and
    ranking information. This network can be trained end-to-end for
    crowd counting. VGG-conv refers to the convolutional layers of the VGG-16 network. See text for more details.}
  \label{fig:framework}
\end{figure*}

\begin{algorithm}[tb]
\centering
\begin{tabularx}{\columnwidth}{rp{0.75\columnwidth}}
\hline
\textbf{Input:}  & A crowd scene image, number of patches $k$ and scale factor $s$. \\
\textbf{Step 1:} & Choose an anchor point randomly from the anchor
region. The anchor region is defined to be $1/r$ the size of the original image, centered at the original image center, and with the same aspect ratio as the original image. \\
\textbf{Step 2:} & Find the largest square patch centered at the
anchor point and contained within the image boundaries. \\
\textbf{Step 3:} & Crop $k-1$ additional square patches, reducing size
iteratively by a scale factor $s$. Keep all patches centered at
anchor point. \\
\textbf{Step 4:} & Resize all $k$ patches to input size of network.\\
\textbf{Output:} & A list of patches ordered according to the number of persons in the patch.\\  \hline
\end{tabularx}
\caption{: Algorithm to generate ranked datasets.}
\label{table:algorithm2}
\end{algorithm}

\section{Learning from ranked image sets} \label{sec:method}
In this section we describe our approach to training a CNN to estimate the number of persons in dense crowd scenes. We use the ranked image set generation technique described in the previous section to generate data for the self-supervised task of ranking crowd images. We first introduce the network architectures used for counting and ranking, and then describe three different approaches to combining both losses.

\subsection{Crowd density estimation network}
Here we explain the network architecture which is trained on available crowd counting datasets with ground truth annotations. This network regresses to a crowd density image which indicates the number of persons per pixel (examples of such maps are given in Figure~\ref{fig:den}). A summation of all values in such a crowd density image gives an estimate of the number of people in the scene. In the experimental section we will consider this network as the baseline method to which we compare.

Our baseline network is derived from the VGG-16 network~\cite{simonyan2014very}, which consists of 13 convolutional layers followed by three fully connected layers. We adapt the network to regress to person density maps. We remove its two fully connected layers, and the max-pooling layer (pool5) to prevent further reduction of spatial resolution. In their place we add a single convolutional layer (a single $3 \times 3 \times 512$ filter with stride 1 and zero padding to maintain same size) which directly regresses to the crowd density map. As the counting loss, $L_c$, we use the Euclidean distance between the estimated and ground truth density maps:
\begin{equation}
\label{eul}
L_c = \frac { 1 }{ M } \sum _{ i=1 }^{ M } (y_{i}-\hat{y}_{i})^2
\end{equation}
where $M$ is the number of images in a training batch, $y_i$ is ground truth person density map of the $i$-th image in the batch, and the prediction from the network as $\hat{y}_i$. The network is indicated in orange in Figure~\ref{fig:framework}. 

Ground truth annotations for crowd counting typically consist of a set of coordinates which indicate the 'center' (typically head center of a person). To convert this data to crowd density maps, we place a Gaussian with standard deviation of 15 pixels and sum these for all persons in the scene to obtain $y_i$. This is a standard procedure and is also used in~\cite{onoro2016towards,zhang2016single}.

The fact that we derive our architecture from the VGG-16 network has the advantage of being able to use pre-trained features from ImageNet. Given the large success of pre-trained features in neural networks, it is somewhat surprising to note that the vast majority of deep learning methods for crowd counting train from scratch~\cite{zhang2016single,Sam_2017_CVPR,sindagi2017generating}. We found, however, that using pre-trained features significantly improves results. 

To further improve the performance of our baseline network, we
introduce multi-scale sampling from the available labeled datasets. Instead of using the whole image as an input,
we randomly sample square patches of varying size (from 56 to 448
pixels). In the experimental section we verify that this multi-scale
sampling is important for good performance. Since we are processing
patches rather than images, we will use $\hat{y}_{i}$ to refer to the estimate of patch $i$ from now on. The importance of multi-scale processing of crowd data was also noted in~\cite{boominathan2016crowdnet}.

\begin{figure}[tpb]
\centering
  \includegraphics[width=\columnwidth]{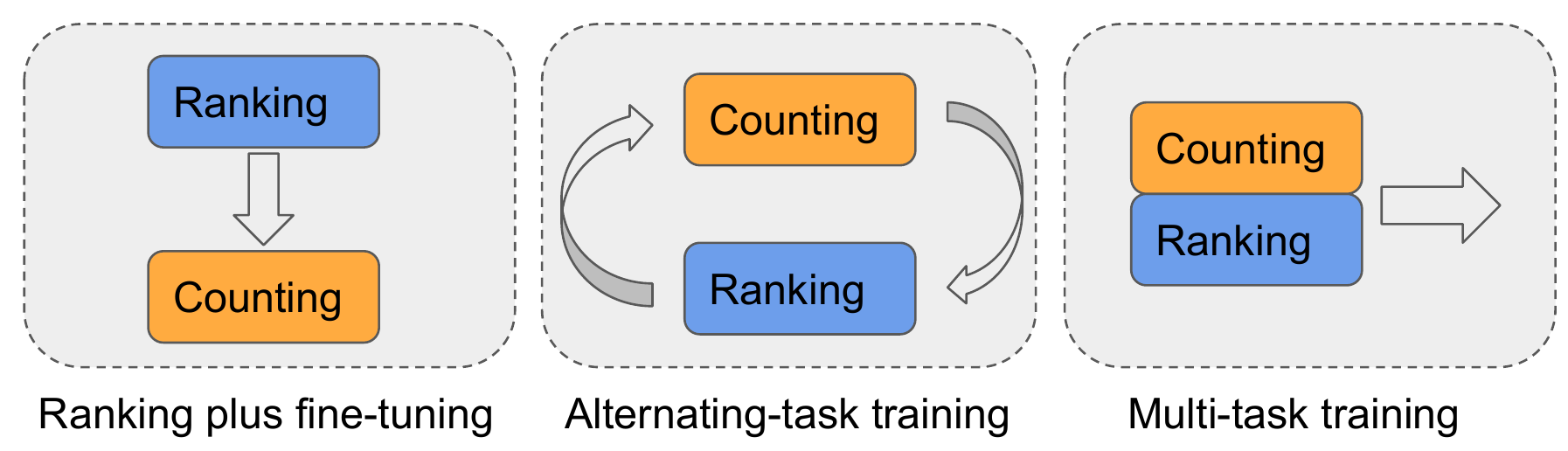}
  \caption{Three ways to combine ranking and counting datasets.}
  \label{fig:three}
\end{figure}

\subsection{Crowd ranking network}
In the previous section we explained how to collect abundantly
available data for crowd counting. This data does not have crowd
density maps and only ranking data is available via the sampling
procedure described in Algorithm~\ref{table:algorithm2}. This ranking
indicates that an equal number or more persons are
present in a patch when compared to another. Here we present the
network which is trained based on this information.
For this purpose, we replace the Euclidean loss by an average pooling
layer followed by a ranking loss (network in blue in
Figure~\ref{fig:framework}). First, the average pooling layer converts
the density map into an estimate of the number of persons per spatial
unit $\hat c(I_i)$ according to:
\begin{equation}
\hat c\left(I_i \right) = \frac{1}{M}\sum\limits_j {\hat y_i \left( {x_j } \right)},
\end{equation}
where $x_j$ are the spatial coordinates of the density map, and
$M=14\times14$ is the number of spatial units in the density map. The
ranking which is on the total number of persons in the patch
$\hat C_i$ also directly holds for its normalized version $\hat c_i$,
since $\hat C\left(I_i \right)=M\times \hat c\left(I_i \right)$.

To enforce the ranking, we apply the pairwise ranking hinge loss which
for a single pair is defined as:
\begin{equation} \label{eq:ranking}
L_r = \max{(0,\hat c \left(I_2\right)- \hat c\left(I_1\right)+\varepsilon )},
\end{equation}
where $\varepsilon$ is the margin, which is set to zero in our
case. This loss increases with the difference between two count
estimates when their order does not respect the correct
ranking. Without loss of generality, we assume that the rank of $\hat
c(I_1)$ is higher than $\hat c(I_2)$.

The gradient with respect to the network parameters $\theta$ of the
loss in Eq.~\ref{eq:ranking} is given by:
\begin{eqnarray}
\nabla_{\theta} L_r \! = \!
\begin{cases}
0 & \!\!\! \mbox{if } \hat c \left(I_2\right) \! - \! \hat c \left(I_1\right) + \varepsilon \le 0 \\
\nabla_{\theta}\hat c \left(I_2\right)-\nabla_{\theta}\hat c \left(I_1\right) & \!\!\! \mbox{otherwise} 
\end{cases}
\end{eqnarray}
When network outputs the correct ranking there is no backpropagated
gradient. However, when the network estimates are not in accordance
with the correct ranking the backpropagated gradient causes the
network to \emph{increase} its estimate for the patch with lower score
and to \emph{decrease} its estimate for the one with higher score
(note that in backpropagation the gradient is subtracted).

A typical implementation of the ranking loss would involve a Siamese network~\cite{chopra2005learning} where two images are sent through parallel branches of the network which share their parameters. However, in~\cite{liu2017rankiqa} the authors show that it is computationally advantageous (and sometimes leads to better minima) to send the images in a batch through a single branch and combine them when computing the ranking loss. The ranking loss is then computed with:
\begin{equation} \label{eq:ranking-fast}
L_r = \sum _{ i=1 }^{ M }\sum _{j \in S\left(i\right)} \max{(0,\hat c \left(I_j\right)-\hat c\left(I_i\right)+\varepsilon )}
\end{equation}
where $S\left(i\right)$ is the set of patches containing fewer people than patch $i$. Note that this relation is only defined for patches which are contained by patch $i$. In practice we sample minibatches of 25 images which contain 5 sets of 5 images which can be compared among them resulting in a total of $5\times\left(4+3+2+1\right)=50$ pairs in one minibatch.

\begin{table*}
\centering
\begin{tabular}{lccccc|c|c}
\hline
\textbf{Method}          & \textbf{Split 1} & \textbf{Split 2} & \textbf{Split 3} & \textbf{Split 4} & \textbf{Split 5} & \textbf{Ave MAE}  & \textbf{Ave MSE}      \\ \hline \hline
Basic CNN       & 701.41 & 394.52 & 497.57 & 263.56 & 415.23 & 454.45 & 620.95  \\
\: + Pre-trained model       & 570.01 & 350.63 & 334.89 & 184.79 & 202.41 & 328.54 & 443.38  \\
\: + multi-scale       & 532.85 & 307.43 & 266.75 & 216.96 & 216.35 & 308.06 & 408.70 \\
 Ranking plus fine-tuning       & 552.68 & 375.38 &  241.28 & 211.66 & 247.70 & 325.73 &429.28 \\
 Alternating-task training & 454.33 & 350.63 & 172.52 & 214.03 &235.70&  285.44 &\textbf{401.29} \\
 Multi-task training & 443.68 & 340.31 & 196.76 & 218.48 & 199.54 & \textbf{279.60 }  &408.07 \\ \hline

\end{tabular}
\caption{MAE and MSE on the UCF\_CC\_50 dataset with five-fold
  cross validation. The Basic CNN is trained from scratch on the
  training set. The second row is the VGG-16 network fine-tuned
  starting from a pre-trained ImageNet model. The third row is the
  VGG-16 network trained with multi-scale data augmentation and starting from a
  pre-trained model. Results for combining both ranking and counting
  datasets are in the last three rows.}
\label{table:ablation}

\end{table*}

\subsection{Combining counting and ranking data}
Here we discuss three approaches to combining ground truth labeled
crowd scenes with data for which only rank information is
available. These three approaches are depicted in
Figure~\ref{fig:three}. We shortly introduce them here:

\begin{itemize}
\item \textbf{Ranking plus fine-tuning:} In this approach the network
  is first trained on the large dataset of ranking data, and is next
  fine-tuned on the smaller dataset for which density maps are
  available. To the best of our knowledge this is the approach which
  is used by all self-supervised methods in
  vision~\cite{doersch2015unsupervised,pathak2016context,zhang2016colorful,noroozi2017representation,liu2017rankiqa}.
  
\item \textbf{Alternating-task training:} While ranking plus
  fine-tuning works well when the two tasks are closely related,
  it might perform bad for crowd counting because no supervision is
  performed to indicate what the network is actually supposed to
  count. Therefore, we propose to \emph{alternate} between the tasks
  of counting and ranking. In practice we perform train for 300
  mini-batches on a single task before switching to the other, then
  repeat.  
  
\item \textbf{Multi-task training:} In the third approach, we add the
  self-supervised task as a proxy to the supervised counting task and
  train both simultaneously.  In each minibatch we sample data from
  both ranking and labeled datasets and train both tasks
  simultaneously as shown in Figure~\ref{fig:framework}. The loss
  function for multi-task training is:
\begin{equation}
L\quad =\quad { L }_{ c }\quad +\quad { \lambda L }_{ r },
\end{equation}
where $\lambda$ sets the relative weight between the counting and
ranking loss. 

\end{itemize}
In the next section we compare these three approaches on several
standard dataset for crowd counting.

\section{Experiments} \label{sec:experiments}

We report on several experiments to evaluate our
approach with respect to baselines and the state-of-the-art
methods.\footnote{Code and models: \href{url}{https://github.com/xialeiliu/CrowdCountingCVPR18}}

\subsection{Datasets and Experimental Protocol}
We use two standard benchmark crowd counting datasets. The
\emph{UCF\_CC\_50} dataset is a very challenging dataset introduced
by~\cite{idrees2013multi}. It contains 50 annotated images of
different resolutions, illuminations and scenes. The variation of
densities is very large among images from 94 to 4543 persons with
an average of 1280 persons per image.

The \emph{ShanghaiTech} dataset introduced by~\cite{zhang2016single}
is a large-scale crowd counting dataset consisting of 1198 images with
330,165 annotated heads. This dataset includes two parts: 482 images
in Part\_A which are randomly crawled from the Internet, and 716
images in Part\_B which are taken from busy streets. Both parts are
further divided into training and evaluation sets. The training and
test of Part\_A has 300 and 182 images, respectively, whereas that of
Part\_B has 400 and 316 images, respectively.

Following existing work, we use the mean absolute
error (MAE) and the mean squared error (MSE) to evaluate different
methods. These are defined as follows:
\begin{equation}
\begin{aligned} 
MAE\quad =\quad \frac { 1 }{ N } \sum _{ i=1 }^{ N }{ \left| { C \left(I_i\right) }-{ \hat  C \left(I_i\right) } \right| \quad ,} \\ MSE\quad =\quad \sqrt { \frac { 1 }{ N } \sum _{ i=1 }^{ N }{ { \left( C \left(I_i\right)-{ \hat C \left(I_i\right)  } \right)  }^{ 2 } }  } 
\end{aligned}
\end{equation}
where $N$ is the number of test images, $C \left(I_i\right)$ is the
ground truth number of persons in the $i$th image and
$\hat C \left(I_i\right)$ is the predicted number of persons in the
$i$th image.

We use the Caffe~\cite{jia2014caffe} framework and train using
minibatch Stochastic Gradient Descent (SGD). The minibatch size for
both ranking and counting is 25, and for multi-task training is
50. For the ranking plus fine-tuning method, the learning rate is 1e-6
for both ranking and fine-tuning. Similarly, for the alternating-task
training method, the steps for training both tasks are 300
iterations. For the multi-task method, we found $\lambda=100$ to provide good results on all datasets. Learning rates are decreased by a factor
of 0.1 every 10K iterations for a total of 20K iterations. For both
training phases we use $\ell_2$ weight decay (set to 5e-4). During
training we sample one sub-image from each training image per
epoch. We perform down-sampling of three scales and up-sampling of one
scale on the UCF\_CC\_50 dataset and only up-sampling of one scale is
used on the ShanghaiTech dataset. The number of ranked crops $k=5$, the scale factor $s=0.75$ and the anchor region $r=8$.

\subsection{Ablation study}
We begin with an ablation study on the UCF\_CC\_50 dataset. The aim is
to evaluate the relative gain of the proposed improvements and
to evaluate the use of a ranking loss against the baseline. The ranked
images in this experiment are generated from the Keyword dataset. The results are summarized in Table~\ref{table:ablation}. We can observe the benefit of using a pre-trained ImageNet model in crowd counting, with a significant drop in MAE of around 28\% compared to the model trained from scratch. By using both multi-scale data augmentation and starting from a pre-trained model, another improvement of around 6\% is obtained.

Next, we compare the three methods we propose for combining the
ranking and counting losses. The ``Ranking plus fine-tuning'' method,
which is the approach applied by all self-supervised methods in the
literature~\cite{doersch2015unsupervised,pathak2016context,zhang2016colorful,noroozi2017representation,liu2017rankiqa},
performs worse than directly fine-tuning from a pre-trained ImageNet
model. This is probably caused by the poorly-defined nature of the
self-supervised task. To optimize this task the network could decide
to count anything, e.g. `hats', `trees', or `people',
all of which would agree with the ranking constraints that are
imposed. By jointly learning both the self-supervised and crowd
counting tasks, the self-supervised task is forced to focus on
counting persons. As a result the ``Alternating-task training'' method
improves the MAE by about 12\% when compared to the direct fine-tuning
method. Moreover, the ``Multi-task training'' approach reduces the MAE
further to 279.6. Given its excellent results we consider only the
``Multi-task training'' approach for the remainder of the experiments.

We also probe how performance scales with increasing training data. We
ran an experiment in which we incrementally add supervised training
data from Part\_A of the ShanghaiTech data. Our approach, using only
60\% of the labeled data, yields about the same accuracy as training
the counting objective alone on 100\% of this data.

\begin{table}
\centering
\resizebox{0.9\columnwidth}{!}{%
\begin{tabular}{rcc}
\hline
\textbf{Method} & \textbf{MAE}   & \textbf{MSE}   \\ \hline \hline
Idrees et al.  \cite{idrees2013multi}     & 419.5 & 541.6 \\
Zhang et al. (2015) \cite{zhang2015cross}       & 467.0 & 498.5 \\
Zhang et al. (2016) \cite{zhang2016single}      & 377.6 & 509.1 \\
Onoro et al.  \cite{onoro2016towards}      & 333.7 & 425.2 \\
Walach et al. \cite{walach2016learning}     & 364.4 & 341.4 \\
Babu Sam et al. \cite{Sam_2017_CVPR}     & 318.1 & 439.2 \\
Sindagi et al. \cite{sindagi2017generating}       & 295.8 & \textbf{320.9} \\ \hline
\textbf{Ours}: Multi-task (Query-by-example)   & 291.5 & 397.6 \\
\textbf{Ours}: Multi-task (Keyword)   & \textbf{279.6} & 388.9 \\ \hline
\end{tabular}}
\caption{MAE and MSE error on the UCF\_CC\_50 dataset.}
\label{table:ucf}
\end{table}

\begin{table}
\centering
\resizebox{0.9\columnwidth}{!}{%
\begin{tabular}{r|rr|rr}
\hline
       & \multicolumn{2}{c|}{\textbf{Part A}} & \multicolumn{2}{c}{\textbf{Part B}} \\
\textbf{Method} & \textbf{MAE}          & \textbf{MSE}         & \textbf{MAE}          & \textbf{MSE}         \\ \hline \hline
Zhang et al. (2015) \cite{zhang2015cross}     & 181.8        & 277.7       & 32.0         & 49.8        \\
Zhang et al. (2016) \cite{zhang2016single}     & 110.2        & 173.2       & 26.4         & 41.3        \\
Babu Sam et al. \cite{Sam_2017_CVPR}    & 90.4         & 135.0       & 21.6         & 33.4        \\
Sindagi et al. \cite{sindagi2017generating}      & 73.6         & \textbf{106.4}       & 20.1         & 30.1        \\ \hline
\textbf{Ours}: Multi-task (Query-by-example)   & \textbf{72.0}        & 106.6 & 14.4         & 23.8      \\
\textbf{Ours}: Multi-task (Keyword)   & 73.6         & 112.0      & \textbf{13.7}         & \textbf{21.4}        \\ \hline
\end{tabular}}
\caption{MAE and MSE error on the ShanghaiTech dataset.}
\label{table:shanghai}
\end{table}

\subsection{Comparison with the state-of-the-art}

\begin{figure*}[tpb]
\centering

\includegraphics[width=0.30\textwidth]{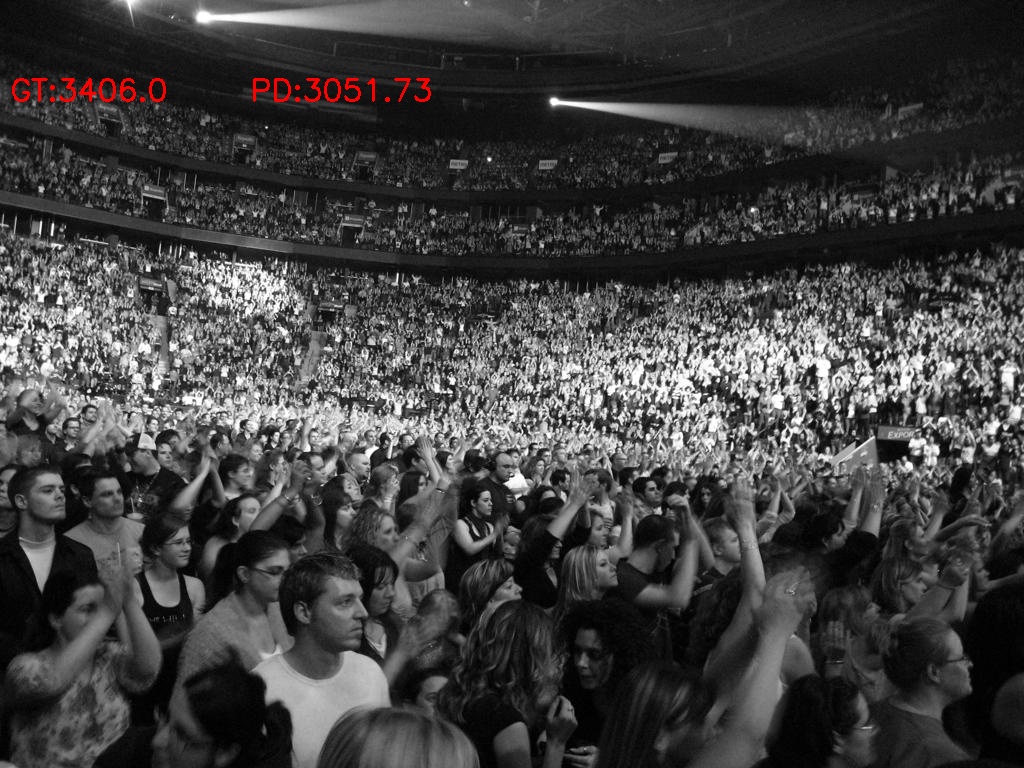} 
 \includegraphics[width=0.30\textwidth]{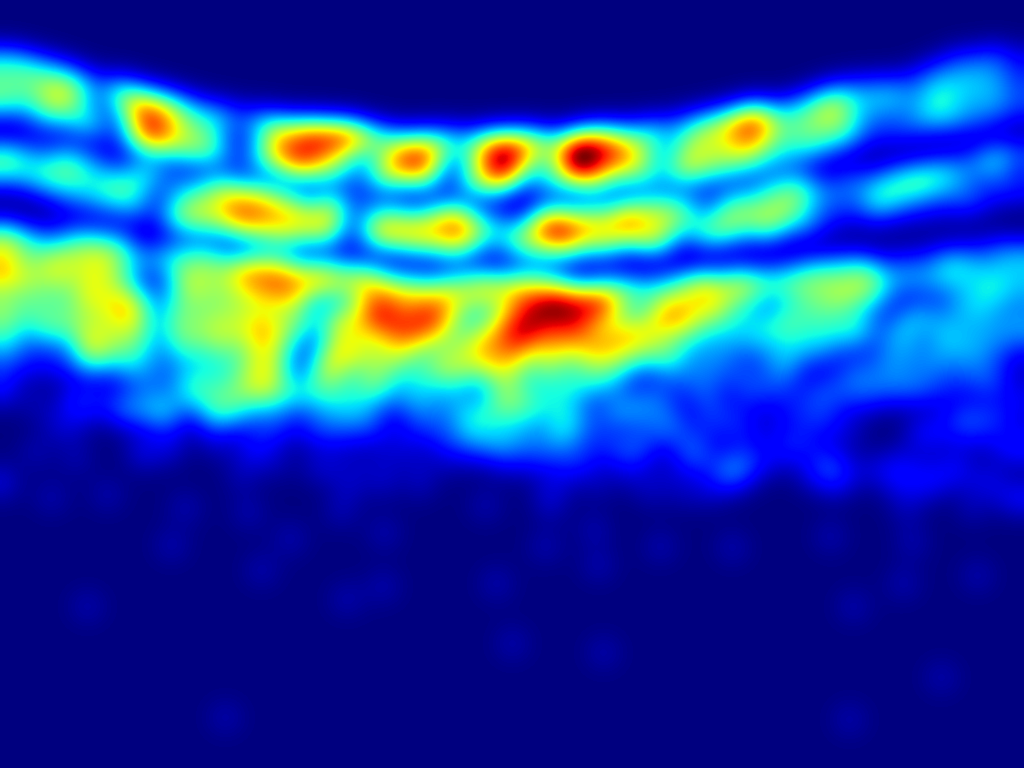}
 \includegraphics[width=0.30\textwidth]{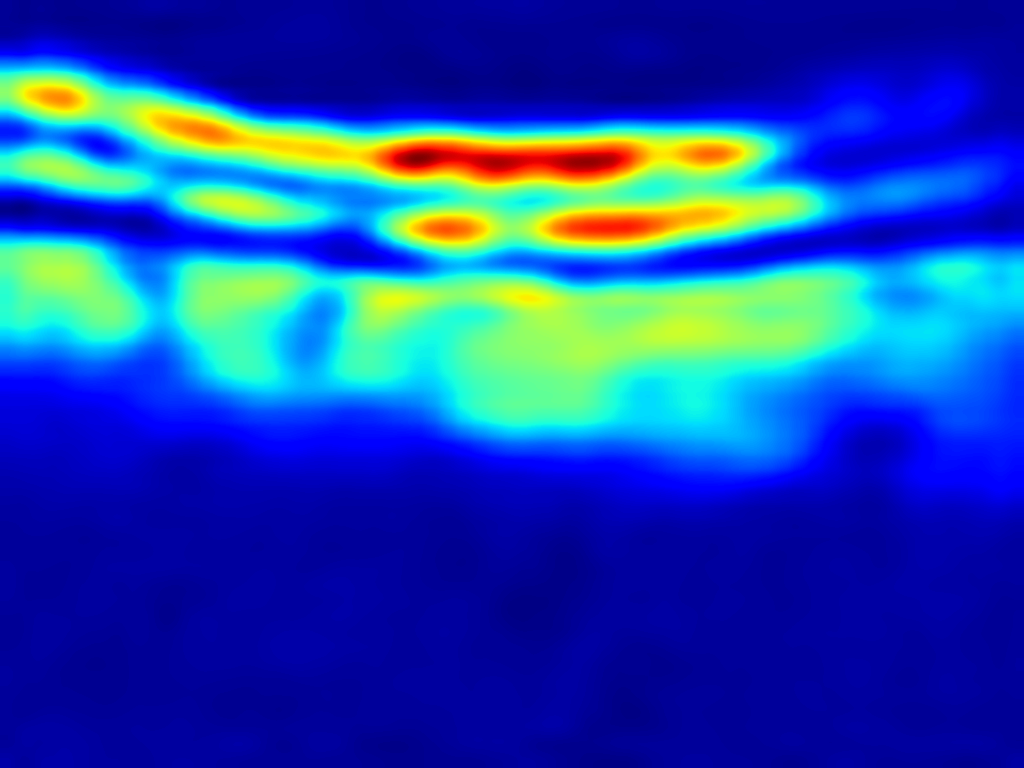}\\
 \includegraphics[width=0.30\textwidth]{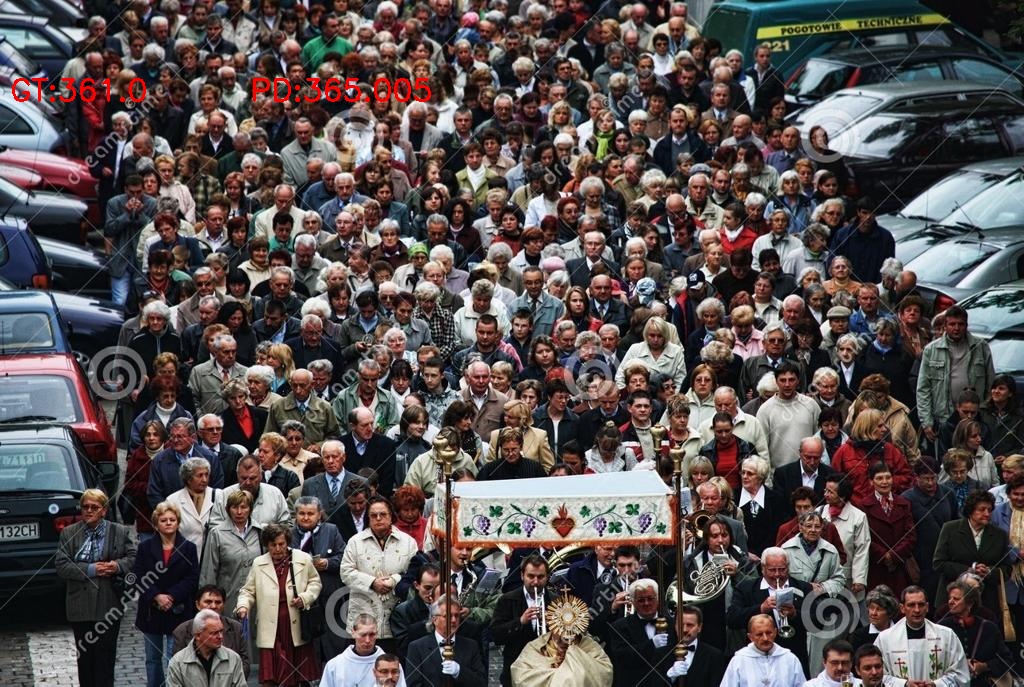}
 \includegraphics[width=0.30\textwidth]{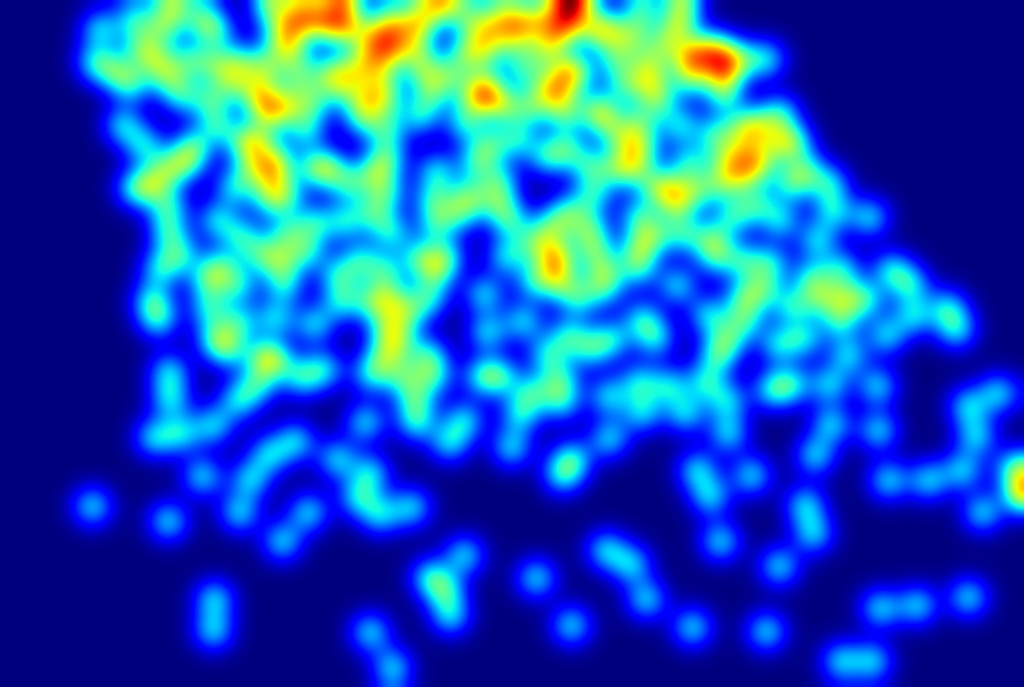}
 \includegraphics[width=0.30\textwidth]{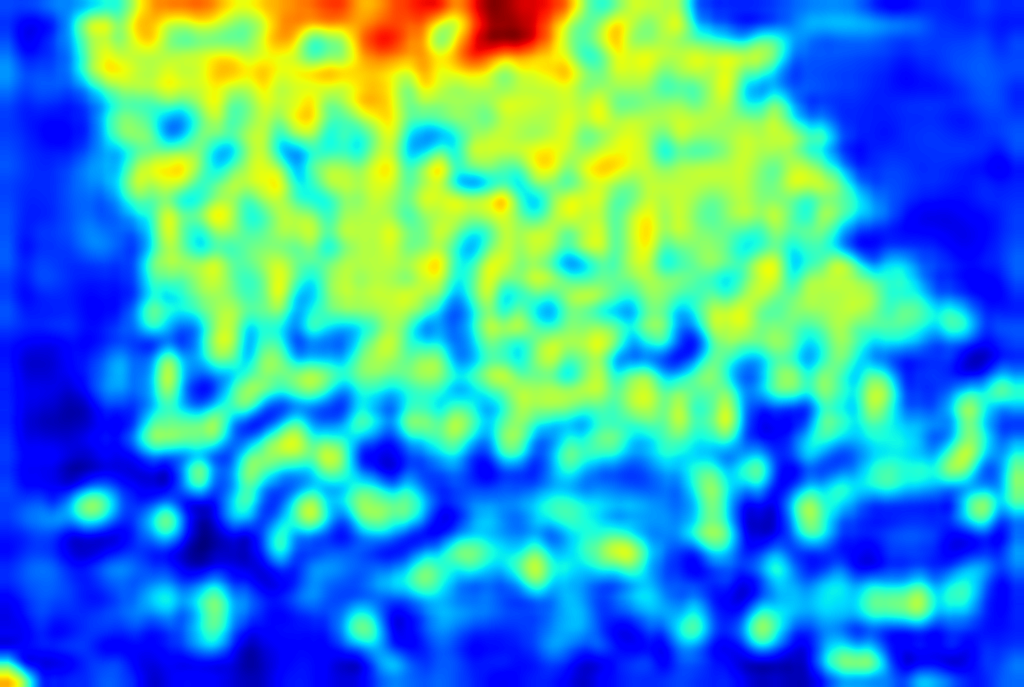}

\caption{Examples of predicted density maps for the UCF\_CC\_50 (Top row, true count: 3406 prediction: 3052) and ShanghaiTech datasets (Bottom row, true count: 361 prediction: 365).  Left column: crowd image. Middle column: ground truth. Right column: prediction.}
  \label{fig:den}
\end{figure*}

We start with the results on the UCF\_CC\_50 dataset.  A five-fold
cross-validation was performed for evaluating the methods. Results are
shown in Table~\ref{table:ucf}. Our multi-task training method with
the Keyword dataset reduces the MAE error from 295.8 to 279.6 compared to
the state-of-the-art. However the MSE of our method on UCF\_CC\_50
dataset is worse then the state-of-the-art
methods~\cite{walach2016learning,sindagi2017generating}, which means
our methods works better in general but has more extreme
outliers. Compared to training on the Keyword dataset, learning from
the Query-by-example dataset is slightly worse, which might be because
most images from UCF\_CC\_50 are black and white with low
resolution, which often does not lead to satisfactory query
results. An example of prediction in UCF\_CC\_50 using our
network is shown in Figure~\ref{fig:den}.

Next we compare with state-of-the-art on the two sets of the
ShanghaiTech dataset.  As shown in Table~\ref{table:shanghai}, similar 
conclusions as on UCF\_CC\_50 can be drawn. We see that using the
our approach further improves by about 2\% on ShanghaiTech. For both Part\_A and Part\_B, our approach surpasses the
state-of-the-art method~\cite{sindagi2017generating}.  An example of
prediction by our network on ShanghaiTech is given in
Figure~\ref{fig:den}. For comparison we also provide the results of
our baseline method (including fine-tuning from a pre-trained model
and multi-scale data augmentation) on this dataset: $MAE=77.7$ and $MSE=115.9$ on
Part A, and $MAE=14.7$ and $MSE=24.7$ on Part B. 
On Part B our baseline already obtains state-of-the-art, with the best
results for the multi-task approach obtaining around a 30\%
improvement when compared to the state-of-the-art. It should also be
noted that the method of~\cite{sindagi2017generating} is complementary
to ours and an approach which combines both methods is expected to
further improve results.

\subsection{Evaluation on transfer learning}

As proposed in~\cite{zhang2016single}, to demonstrate the
generalization of the learned model, we test our method in the
transfer learning setting by using Part\_A of the ShanghaiTech dataset
as the source domain and using UCF\_CC\_50 dataset as the target
domain. The model trained on Part\_A of ShanghaiTech is used to
predict the crowd scene images from UCF\_CC\_50 dataset, and the
results can be seen in Table~\ref{table:transfer}. Using only counting
information improves the MAE by 12\% compared to reported results
in~\cite{zhang2016single}. By combining both ranking and counting
datasets, the MAE decreases from 349.5 to 337.6, and MSE decreases
from 475.7 to 434.3. In conclusion, these results show that our method
significantly outperforms the only other work reporting results
on the task of cross-dataset crowd counting.

\begin{table}[tb]
\centering

\begin{tabular}{rcc}
\hline
\textbf{Method} & \textbf{MAE}   & \textbf{MSE}   \\ \hline \hline
Zhang et al. (2016) \cite{zhang2016single}      & 397.6 & 624.1 \\ \hline
\textbf{Ours}: Counting only        & 349.5 & 475.7 \\
\textbf{Ours}: Multi-task   & 337.6 & 434.3 \\ \hline
\end{tabular}
\caption{Transfer learning across datasets. Models were trained on
  Part\_A of ShanghaiTech and tested on UCF\_CC\_50.}
\label{table:transfer}
\end{table}

\section{Conclusions}
In this work we proposed a method for crowd counting. The main novelty
is based on the observation that a crop which is contained within a
larger crop must contain fewer or an equal number of persons than the
larger crop. This allows us to address one of the main problems for
crowd counting, namely the lack of large training datasets. Our
approach enables the exploitation of abundantly available training
data from the Internet by automatically generating rankings from
them. We showed how this additional data can be leveraged with
available annotated data in a multi-task network. 

Experiments show that the proposed self-supervised task improves
results significantly when compared to a network which is only trained
on the annotated data. We show that incorporating the self-supervised
task in a multi-task approach obtains optimal results. Furthermore, we
obtain state-of-the-art results on two challenging datasets for crowd
counting, namely the ShanghaiTech and the UCF\_CC\_50
dataset. Finally, we show that the learned models generalize well to
other datasets, significantly outperforming the only other crowd
counting method which reports on this transfer learning task.

\minisection{Acknowledgements}
We acknowledge  the  Spanish project TIN2016-79717-R, the CHISTERA project M2CR (PCIN-2015-251) and the CERCA Programme / Generalitat de Catalunya. Xialei Liu acknowledges the Chinese Scholarship Council (CSC) grant No.201506290018. We also acknowledge the generous GPU donation from NVIDIA.

\newpage
{\small
\bibliographystyle{ieee}
\bibliography{refs}

\begin{thebibliography}{10}\itemsep=-1pt

\bibitem{Sam_2017_CVPR}
D.~Babu~Sam, S.~Surya, and R.~Venkatesh~Babu.
\newblock Switching convolutional neural network for crowd counting.
\newblock In {\em CVPR}, 2017.

\bibitem{boominathan2016crowdnet}
L.~Boominathan, S.~S. Kruthiventi, and R.~V. Babu.
\newblock Crowdnet: a deep convolutional network for dense crowd counting.
\newblock In {\em Proc. ACM on Multimedia Conference}, 2016.

\bibitem{chan2008privacy}
A.~B. Chan, Z.-S.~J. Liang, and N.~Vasconcelos.
\newblock Privacy preserving crowd monitoring: Counting people without people
  models or tracking.
\newblock In {\em CVPR}, 2008.

\bibitem{chan2009bayesian}
A.~B. Chan and N.~Vasconcelos.
\newblock Bayesian poisson regression for crowd counting.
\newblock In {\em ICCV}, 2009.

\bibitem{chan2012counting}
A.~B. Chan and N.~Vasconcelos.
\newblock Counting people with low-level features and bayesian regression.
\newblock {\em IEEE Transactions on Image Processing}, 21(4):2160--2177, 2012.

\bibitem{chen2012feature}
K.~Chen, C.~C. Loy, S.~Gong, and T.~Xiang.
\newblock Feature mining for localised crowd counting.
\newblock In {\em BMVC}, 2012.

\bibitem{chen2009ranking}
W.~Chen, T.-Y. Liu, Y.~Lan, Z.-M. Ma, and H.~Li.
\newblock Ranking measures and loss functions in learning to rank.
\newblock In {\em NIPS}, 2009.

\bibitem{chopra2005learning}
S.~Chopra, R.~Hadsell, and Y.~LeCun.
\newblock Learning a similarity metric discriminatively, with application to
  face verification.
\newblock In {\em CVPR}, 2005.

\bibitem{cong2009flow}
Y.~Cong, H.~Gong, S.-C. Zhu, and Y.~Tang.
\newblock Flow mosaicking: Real-time pedestrian counting without scene-specific
  learning.
\newblock In {\em Computer Vision and Pattern Recognition, 2009. CVPR 2009.
  IEEE Conference on}, pages 1093--1100. IEEE, 2009.

\bibitem{doersch2015unsupervised}
C.~Doersch, A.~Gupta, and A.~A. Efros.
\newblock Unsupervised visual representation learning by context prediction.
\newblock In {\em ICCV}, 2015.

\bibitem{dollar2012pedestrian}
P.~Dollar, C.~Wojek, B.~Schiele, and P.~Perona.
\newblock Pedestrian detection: An evaluation of the state of the art.
\newblock {\em IEEE trans. on pattern analysis and machine intelligence},
  34(4):743--761, 2012.

\bibitem{fu2015fast}
M.~Fu, P.~Xu, X.~Li, Q.~Liu, M.~Ye, and C.~Zhu.
\newblock Fast crowd density estimation with convolutional neural networks.
\newblock {\em Engineering Applications of Artificial Intelligence}, 43:81--88,
  2015.

\bibitem{idrees2013multi}
H.~Idrees, I.~Saleemi, C.~Seibert, and M.~Shah.
\newblock Multi-source multi-scale counting in extremely dense crowd images.
\newblock In {\em CVPR}, 2013.

\bibitem{jia2014caffe}
Y.~Jia, E.~Shelhamer, J.~Donahue, S.~Karayev, J.~Long, R.~Girshick,
  S.~Guadarrama, and T.~Darrell.
\newblock Caffe: Convolutional architecture for fast feature embedding.
\newblock {\em arXiv preprint arXiv:1408.5093}, 2014.

\bibitem{kang2016crowd}
D.~Kang, D.~Dhar, and A.~B. Chan.
\newblock Crowd counting by adapting convolutional neural networks with side
  information.
\newblock {\em arXiv preprint arXiv:1611.06748}, 2016.

\bibitem{krizhevsky2012imagenet}
A.~Krizhevsky, I.~Sutskever, and G.~E. Hinton.
\newblock Imagenet classification with deep convolutional neural networks.
\newblock In {\em NIPS}, pages 1097--1105, 2012.

\bibitem{larsson2017colorization}
G.~Larsson, M.~Maire, and G.~Shakhnarovich.
\newblock Colorization as a proxy task for visual understanding.
\newblock In {\em CVPR}, 2017.

\bibitem{lempitsky2010learning}
V.~Lempitsky and A.~Zisserman.
\newblock Learning to count objects in images.
\newblock In {\em NIPS}, pages 1324--1332, 2010.

\bibitem{li2008estimating}
M.~Li, Z.~Zhang, K.~Huang, and T.~Tan.
\newblock Estimating the number of people in crowded scenes by mid based
  foreground segmentation and head-shoulder detection.
\newblock In {\em International Conference on Pattern Recognition}. IEEE, 2008.

\bibitem{liu2017rankiqa}
X.~Liu, J.~van~de Weijer, and A.~D. Bagdanov.
\newblock Rankiqa: Learning from rankings for no-reference image quality
  assessment.
\newblock In {\em ICCV}, 2017.

\bibitem{loy2013crowd}
C.~C. Loy, K.~Chen, S.~Gong, and T.~Xiang.
\newblock Crowd counting and profiling: Methodology and evaluation.
\newblock In {\em Modeling, Simulation and Visual Analysis of Crowds}, pages
  347--382. Springer, 2013.

\bibitem{ma2013crossing}
Z.~Ma and A.~B. Chan.
\newblock Crossing the line: Crowd counting by integer programming with local
  features.
\newblock In {\em Proceedings of the IEEE Conference on Computer Vision and
  Pattern Recognition}, pages 2539--2546, 2013.

\bibitem{marsden2016fully}
M.~Marsden, K.~McGuiness, S.~Little, and N.~E. O'Connor.
\newblock Fully convolutional crowd counting on highly congested scenes.
\newblock {\em arXiv preprint arXiv:1612.00220}, 2016.

\bibitem{noroozi2017representation}
M.~Noroozi, H.~Pirsiavash, and P.~Favaro.
\newblock Representation learning by learning to count.
\newblock In {\em ICCV}, 2017.

\bibitem{onoro2016towards}
D.~Onoro-Rubio and R.~J. L{\'o}pez-Sastre.
\newblock Towards perspective-free object counting with deep learning.
\newblock In {\em ECCV}. Springer, 2016.

\bibitem{pathak2016context}
D.~Pathak, P.~Krahenbuhl, J.~Donahue, T.~Darrell, and A.~A. Efros.
\newblock Context encoders: Feature learning by inpainting.
\newblock In {\em CVPR}, 2016.

\bibitem{test1}
D.~Sculley.
\newblock Large scale learning to rank.
\newblock In {\em NIPS Workshop on Advances in Ranking}, 2009.

\bibitem{shang2016end}
C.~Shang, H.~Ai, and B.~Bai.
\newblock End-to-end crowd counting via joint learning local and global count.
\newblock In {\em International Conference on Image Processing (ICIP)}, 2016.

\bibitem{sheng2016crowd}
B.~Sheng, C.~Shen, G.~Lin, J.~Li, W.~Yang, and C.~Sun.
\newblock Crowd counting via weighted vlad on dense attribute feature maps.
\newblock {\em IEEE Transactions on Circuits and Systems for Video Technology},
  2016.

\bibitem{simonyan2014very}
K.~Simonyan and A.~Zisserman.
\newblock Very deep convolutional networks for large-scale image recognition.
\newblock {\em ICLR}, 2015.

\bibitem{sindagi2017generating}
V.~A. Sindagi and V.~M. Patel.
\newblock Generating high-quality crowd density maps using contextual pyramid
  cnns.
\newblock In {\em ICCV}, 2017.

\bibitem{sindagi2017survey}
V.~A. Sindagi and V.~M. Patel.
\newblock A survey of recent advances in cnn-based single image crowd counting
  and density estimation.
\newblock {\em Pattern Recognition Letters}, 2017.

\bibitem{walach2016learning}
E.~Walach and L.~Wolf.
\newblock Learning to count with cnn boosting.
\newblock In {\em ECCV}. Springer, 2016.

\bibitem{wang2015deep}
C.~Wang, H.~Zhang, L.~Yang, S.~Liu, and X.~Cao.
\newblock Deep people counting in extremely dense crowds.
\newblock In {\em Proceedings of ACM int. conf. on Multimedia}. ACM, 2015.

\bibitem{zhang2015cross}
C.~Zhang, H.~Li, X.~Wang, and X.~Yang.
\newblock Cross-scene crowd counting via deep convolutional neural networks.
\newblock In {\em CVPR}, 2015.

\bibitem{zhang2016colorful}
R.~Zhang, P.~Isola, and A.~A. Efros.
\newblock Colorful image colorization.
\newblock In {\em ECCV}, 2016.

\bibitem{zhang2016single}
Y.~Zhang, D.~Zhou, S.~Chen, S.~Gao, and Y.~Ma.
\newblock Single-image crowd counting via multi-column convolutional neural
  network.
\newblock In {\em CVPR}, 2016.

\end{thebibliography}
}

\end{document}